%% file: main.tex
\documentclass[sigconf]{acmart}
\input{tools}

\copyrightyear{2024}
\acmYear{2024}
\setcopyright{acmlicensed}
\acmConference[WSDM '24] {Proceedings of the 17th ACM International Conference on Web Search and Data Mining}{March 4--8, 2024}{Merida, Mexico.}
\acmBooktitle{Proceedings of the 17th ACM International Conference on Web Search and Data Mining (WSDM '24), March 4--8, 2024, Merida, Mexico}
\acmPrice{15.00}
\acmISBN{979-8-4007-0371-3/24/03}
\acmDOI{10.1145/3616855.3635802}

\begin{document}

\title{Source Free Graph Unsupervised Domain Adaptation}


\author{Haitao Mao}\authornote{Work done while the author was on internship at Microsoft Research Asia.}
\affiliation{
  \institution{Michigan State University}
  \country{United States}
}
\email{haitaoma@msu.edu}
\author{Lun Du}\authornote{Corresponding author}
\affiliation{
  \institution{Microsoft Research Asia}
  \country{China}
}
\email{lun.du@microsoft.com}
\author{Yujia Zheng}
\affiliation{
  \institution{Carnegie Mellon University}
  \country{United States}
}
\email{yujiazh@cmu.edu}
\author{Qiang Fu}
\affiliation{
  \institution{Microsoft Research Asia}
  \country{China}
}
\email{qifu@microsoft.com}
\author{Zelin Li$^*$}
\affiliation{
  \institution{Microsoft Research Asia}
  \country{China}
}
\email{v-zelinli@microsoft.com}
\author{Xu Chen}
\affiliation{
  \institution{Microsoft Research Asia}
  \country{China}
}
\email{xu.chen@microsoft.com}
\author{Shi Han}
\affiliation{
  \institution{Microsoft Research Asia}
  \country{China}
}
\email{shihan@microsoft.com}
\author{Dongmei Zhang}
\affiliation{
  \institution{Microsoft Research Asia}
  \country{China}
}
\email{dongmeiz@microsoft.com}









\begin{abstract}
Graph Neural Networks (GNNs) have achieved great success on a variety of tasks with graph-structural data, among which node classification is an essential one. Unsupervised Graph Domain Adaptation (UGDA) shows its practical value of reducing the labeling cost for node classification. It leverages knowledge from a labeled graph (i.e., source domain) to tackle the same task on another unlabeled graph (i.e., target domain). Most existing UGDA methods heavily rely on the labeled graph in the source domain. They utilize labels from the source domain as the supervision signal and are jointly trained on both the source graph and the target graph. However, in some real-world scenarios, the source graph is inaccessible because of privacy issues. Therefore, we propose a novel scenario named Source Free Unsupervised Graph Domain Adaptation (SFUGDA). In this scenario, the only information we can leverage from the source domain is the well-trained source model, without any exposure to the source graph and its labels. As a result, existing UGDA methods are not feasible anymore. To address the non-trivial adaptation challenges in this practical scenario, we propose a model-agnostic algorithm called SOGA for domain adaptation to fully exploit the discriminative ability of the source model while preserving the consistency of structural proximity on the target graph. We prove the effectiveness of the proposed algorithm both theoretically and empirically. The experimental results on four cross-domain tasks show consistent improvements in the Macro-F1 score and Macro-AUC.
\end{abstract}






\maketitle
\input{src/introduction}
\input{src/related_work}

\input{src/model}
\input{src/experiment}

\input{src/conclusion}

\bibliographystyle{ACM-Reference-Format}
\balance
\bibliography{main}

\appendix
\input{src/appendix}

\end{document}

%% file: tools.tex
\usepackage{enumitem}
\usepackage{makecell}
\usepackage{xspace}
\usepackage{amsmath,amssymb,amsfonts}
\usepackage{graphicx}
\usepackage{textcomp}
\usepackage{subfigure}
\usepackage{tcolorbox}
\usepackage{booktabs}
\usepackage{tabularx}
\usepackage{lipsum}
\usepackage{multirow}
\usepackage{color, xcolor}

\newcommand{\model}{SOGA}

\usepackage{amssymb}
\newtheorem{defn}{Definition}

\newcommand{\im}{{\it IM}}
\newcommand{\scim}{{\it SC}}

\newtheorem{lem}{Lemma}

%% file: src/introduction.tex
\section{Introduction}
Node classification \cite{kipf2016semi} is a crucial task on graph-structural data such as transaction network \cite{zhang2023company}, citation networks \cite{tang2008arnetminer,song2022learning,mao2023revisiting},  and so on.
Recently, Graph Neural Networks\cite{kipf2016semi,mao2023demystifying} have greatly advanced the performance of node classification. 
However, most existing studies only concentrate on how to classify well on one given graph of a specific domain, while ignoring its performance degradation when applying it to graphs from other domains due to the domain gap. 
For example, regarding two real-world citation networks with papers as nodes and edges representing their citations,
papers published between 2000 - 2010 and papers published between 2010 - 2020 may have significant differences from the following two aspects: (1) Feature distribution shifts as the advanced research topics and high-frequency keywords change over time. 
(2) Discrepancy between graph structures: due to the great success of Deep Learning, papers on machine learning and neuro-science have been more frequently cited in recent years. 
\begin{figure}
    \includegraphics[width=0.4\textwidth]{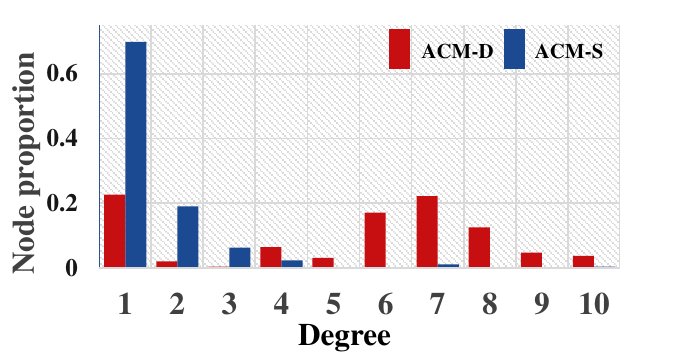}
    \caption{Discrepancy between degree distributions on ACM-D and ACM-S datasets.}
    \label{fig:degree_example}
\end{figure}
More concretely, we select two datasets: ACM-D, and ACM-S, two subgraphs from ACM dataset \cite{yang2020unsupervised} to study their graph structure discrepancy. Their degree distributions are shown in Fig. \ref{fig:degree_example}. It is obvious to see that the node degree distributions on different graphs are different. 
The above problems, i.e., feature distribution shift and graph structure discrepancy, lead to unsatisfactory performance when transferring the GNN model across graphs from different domains to handle the same task.
The naive way to achieve good results on the target graph from a different domain is to label the graph manually and retrain a new model from scratch, which is expensive and time-consuming.
To solve this problem, Unsupervised Domain Adaptation (UDA), a transfer learning technique leveraging knowledge learned from a sufficiently labeled source domain to enhance the performance on an unlabeled target domain, has raised increasing attention recently.

UDA has shown great success on image data \cite{tzeng2014deep, ganin2015unsupervised, saito2017asymmetric} and text data \cite{jiang2007instance, dai2007co}. 
Recently, Unsupervised Graph Domain Adaptation (UGDA) has been proposed as a new application of UDA on graph data. It utilizes important properties of graphs, especially the structural information indicating the correlation between nodes. 
Generally speaking, most existing UGDA methods \cite{yang2020domain, shen2020network, zhang2019dane, wu2020unsupervised} utilize a joint learning framework: 
(1) A feature encoder is trained to align the feature distributions between the source domain and target domain to mitigate the domain gap. 
(2) A classifier is trained on encoded features with cross-entropy loss, supervised by source labels.
The model can achieve satisfying performance with strong discriminative ability on the aligned feature distribution.

However, a crucial requirement for these joint learning methods is access permission to the source data, which might be problematic for both accessibility and privacy issues \cite{voigt2017eu}. 
In the real-world scenario, access to the source domain is not always available (e.g., domain adaptation between two different platforms). 
The usage of sensitive attributes on graphs may lead to potential data leakage and other severe privacy issues.
Therefore, we propose a new scenario, \textbf{Source Free Unsupervised Graph Domain Adaptation} (SFUGDA), in which only the unlabeled target data and the GNN model trained from source data are available for adaptation.

The key challenges in this scenario are two-fold: 
(1) How the model can adapt well to the shifted target data distribution without accessing the source graph for aligning the feature distributions.
(2) How to enhance the discriminative ability of the source model without accessing source labels for supervision.
In this paper, we propose \model{}, a model agnostic \textbf{SO}urce free domain \textbf{G}raph \textbf{A}daptation algorithm, which enables the GNN model trained on the labeled source graph to perform well on the unlabeled target graph.

\model{} addresses these challenges by the following two components: 
(1) Structure Consistency (SC) optimization objective: inspired by the unsupervised graph embedding methods \cite{tzeng2014deep, ribeiro2017struc2vec}, which learn node representations by preserving various graph properties using well-designed objective functions, we propose SC objective to tune the source model to reflect the target graph structure in the model output representation space. It can adapt the source model to the shifted target data distribution.
(2) Information Maximization (IM) optimization objective is proposed to enhance the discriminative ability of the source model by maximizing the mutual information between the target graph and its corresponding output. 
We theoretically prove that IM can improve the confidence of prediction and raise the lower bound of the AUC metric.

Moreover, we usually can neither determine the source model architecture nor its training procedure in practice, other than no accessibility to the source data. 
Our algorithm is also model agnostic which be easily adopted to arbitrary GNN models. 
With this property, our \model{} can easily satisfy the above practical requirements.

In summary, the main contributions of our work are as follows:
\begin{itemize}
    \item We first articulate a new scenario called SFUGDA when we have no access to the source graph and its labels.
    To the best of our knowledge, this is the first work in SFUGDA. 
    \item We propose a model agnostic unsupervised algorithm called \model{} to tackle challenges in SFUGDA. 
    It can both adapt the source model to the shifted target distribution and enhance its discriminative ability with a theoretical guarantee.
    \item Extensive experiments are conducted on real-world datasets. 
    Our \model{} outperforms all the baselines on four cross-domain tasks. 
    Moreover, experimental results verify the model agnostic property as \model{} can be applied with different representative GNN models successfully. 
\end{itemize}

%% file: src/related_work.tex
\section{Related Work}
\subsection{Comparison with topics on Domain Adaptation.}

Unsupervised Graph Domain Adaptation (UGDA) aims to transfer the knowledge learned on a labeled graph from the source domain to an unlabeled graph from the target domain tackling the same task. 
Most existing UGDA methods aim to mitigate the domain gap by aligning the source feature distribution and the target one. 
According to different alignment approaches, they can be roughly divided into two categories. 
(1) Distance-based methods like \cite{shen2020network, yang2020domain} incorporate maximum mean discrepancy (MMD) \cite{borgwardt2006integrating} as a domain distance loss to match the distribution statistical moments at different order.
(2) Domain adversarial methods \cite{zhang2019dane, wu2020unsupervised}
follow the guidance of adversarial training, which confuses generated features across the source domain and the target one to mitigate the domain discrepancy. 
DANE \cite{zhang2019dane} adds an adversarial regularizer inspired by LSGAN \cite{mao2017least}, while UDAGCN \cite{wu2020unsupervised} uses Gradient Reversal Layer \cite{ganin2015unsupervised} and domain adversarial loss to extract cross-domain node embedding.

However, all the above UGDA methods heavily rely on access to the source data, which leads to failure in the SFUGDA scenario where source data is not available anymore.

In computer vision, Source Free Unsupervised Domain Adaptation is a new research task with practical value.
Most existing studies focus on different strategies to generate pseudo labels on
images inspired by \cite{saito2017asymmetric}.
\cite{liang2020we} utilizes the Deep Cluster algorithm to assign cleaner pseudo labels with a global view and an Information
Maximization algorithm to minimize the prediction uncertainty PrDA \cite{kim2020progressive} uses a set-to-set distance to filter confident pseudo labels.
\cite{li2020model} focuses on how to adapt the feature distribution on the target domain by generating similar feature samples from a GAN-based model. 
\cite{yang2021exploiting} encourages label consistency on local affinity neighborhoods based on the key observation that the target data can still form clear data clusters.
However, the above methods designed for image data are not suitable for graph-structural data. Since graph node samples are naturally structured by dependencies (i.e.,
edges) between nodes, strategies focusing on i.i.d. data like images, cannot be well adapted. For example, even when feature distribution stays the same, the graph may still suffer from domain gaps for various structure patterns. Thus, methods for SFUGDA should handle structural dependencies well. Despite the graph structure leading to new challenges, various properties of the graph structure, such as structure proximity, could help the adaptation if modeled properly.

Moreover, our proposed SFUGDA method is more practical than methods for images as most of them need specific-designed source model architecture. They utilize different components like BatchNorm or WeightNorm to implicitly memorize the knowledge from source data. However, it seems not feasible in practice to retrain a specific source model for adaptation. Contrastively, our SOGA can combine with any GNN model with no need for a specific design.

\subsection{Comparison between SFUGDA with other topics on graphs.\label{app:related}}
Graph self-supervised learning~\cite{velickovic2019deep, zhu2020deep, zhang2021canonical, hu2019strategies, hu2020gpt, qiu2020gcc} is a new scenario on the graph which also utilizes the two-stage procedure similar with SFUGDA. 
Typically, those methods will first have an unsupervised learning procedure by creating graph-specific pretext tasks and training on the unlabeled data. This procedure aims to learn a good representation that could benefit different downstream tasks.
Then a supervised fine-tuning procedure is employed with the labeled data for the specific downstream tasks.
However, those methods are not applicable in the SFUGDA scenario as there is no label information in the target domain for the supervised fine-tuning.

More recently, self-supervised learning is also utilized as an unsupervised fine-tuning strategy~\cite{chen2022contrastive, mummadi2021test, wang2020tent}. They typically update the original model by minimizing a self-supervised loss on the target distribution. 
However, most are specifically designed for image data and may not be suitable for graph-structured data. 
\cite{jin2022empowering} is the first to utilize self-supervised learning for finetuning in the graph domain. 
It focuses on learning an adaptive graph transformation from a data-centric perspective rather than learning a well-performed model.

Graph Federated learning is a distributed machine learning approach for privacy which has raised great interest in graph \cite{lalitha2019peer, xie2021federated}. They aggregate a server-side model from multiple decentralized edge devices without data leakage, offering a privacy-preserving mechanism. However, it requires each local data with labeled information. It fails to address the SFUGDA scenario where label information is only available in one single source domain.

%% file: src/model.tex
\section{Preliminary}
\begin{defn}[\textbf{\emph{Node Classification}}]
\label{def:node_classification}
	Node classification is a task to learn a conditional probability $\mathbf{\mathcal{Q}}(\mathbf{y}|\mathbf{G}; \Theta)$ to distinguish the category of each unlabeled node on a \textbf{single} graph $\mathbf{G} = ( \mathbf{V}, \mathbf{E}, \mathbf{X}, \mathbf{y} )$, where $\Theta$ is the model parameters. $\mathbf{V} = \{ v_1, \cdots v_n \}$ is the node set with $n$ nodes and $\mathbf{E}$ is the edge set. 
    $\mathbf{X} \in \mathbb{R}^{n \times d}$ is the node feature matrix, $\mathbf{y} \in \mathbb{R}^n$ is the partially observed node label set of which each element satisfies $y_i \in \left \{1, 2, \cdots k, -1 \right\}$. $y_i = -1$ indicates $i$-th node is unlabeled, $d$ is the feature dimension, and $k$ is the number of categories.
\end{defn}
Node classification differs from typical classification tasks since the latter usually assumes that different samples are i.i.d (independent identical distribution), whereas samples in the former case are correlated through edges.
In deep learning, we use Graph Neural Networks (GNNs) \cite{hamilton2017inductive} to model $\mathbf{\mathcal{Q}}(\mathbf{y}|\mathbf{G}; \Theta)$, the conditional probability to distinguish the category of all nodes, for capturing such relationships. Based on the assumption of localization \cite{defferrard2016convolutional}, the predicted conditional distribution is usually decomposed as follows,  
$\mathbf{\mathcal{Q}}(\mathbf{y}|\mathbf{G}; \Theta) = \prod_{v_i \in \mathbf{V}} q(y_i|x_i, \mathbf{\mathcal{N}_i}; \Theta)$, where $q(y_i|x_i, \mathbf{\mathcal{N}_i}; \Theta)$ is the conditional probability to distinguish the category of one single node $v_i$. 
Notice that, $\mathbf{G}$ includes the feature $x_i$ and the neighbor information $\mathbf{\mathcal{N}_i}$ for each node $v_i$. Following the GNN model, Cross Entropy is usually adopted as the loss function:
\begin{equation}
\label{equ:CE}
\mathbb{E}_{v_i \sim p(v)} \left[-\sum_{y = 1}^{k} p(y|x_i, \mathbf{\mathcal{N}_i}) \log q(y | x_i, \mathbf{\mathcal{N}_i}; \Theta)\right],
\end{equation} 
where $p(v)$ is the prior distribution and $p(y|x, \mathcal{N})$ is the oracle conditional distribution.
As information in the node $v_i$ contains its own feature $x_i$ and neighborhood information $\mathbf{\mathcal{N}_i}$, we will simplify $p(x_i, \mathbf{\mathcal{N}_i})$ to $p(v_i)$ for brevity.  
 
According to Def. \ref{def:node_classification}, a typical node classification task is defined on a single graph with partial supervision. To better leverage the knowledge from a labeled graph (namely source graph) to tackle the same node classification task on another unlabeled graph (namely target graph), Unsupervised Graph Domain Adaptation (UGDA) \cite{zhang2019dane} is proposed.
We then give a clear definition of UGDA which has already been well recognized in \cite{shen2020adversarial, zhang2019dane, wu2020unsupervised, jin2020active, shen2020adversarial}. 
\begin{figure}[ht]
    \centering
    \includegraphics[width=0.48\textwidth]{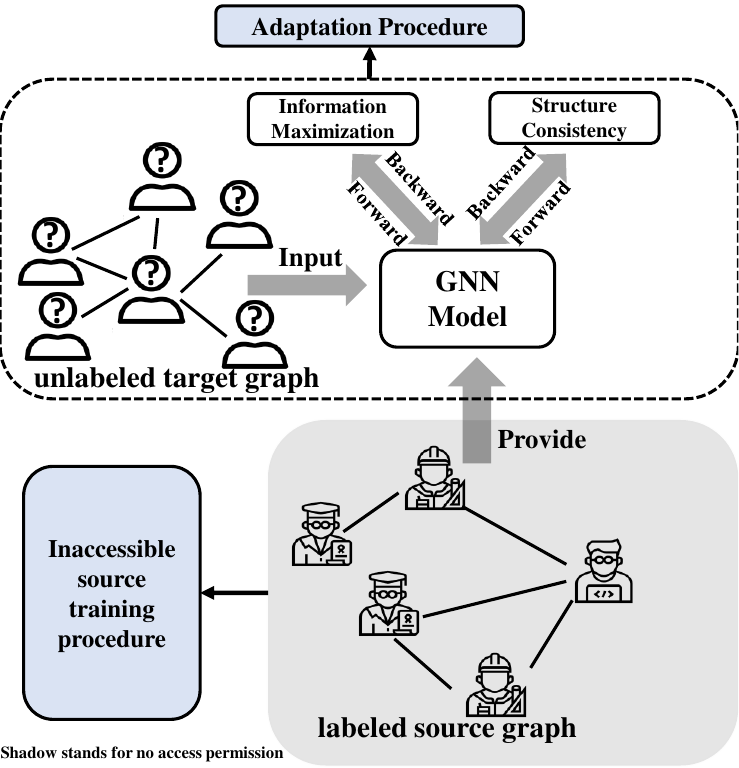}
    \caption{The detailed procedure of the \model{} algorithm. 
    In the SFUGDA scenario, the source training procedure and labeled source graph in the shadow box are not accessible.
    Our algorithm only includes the upper dashed box describing the adaptation procedure. 
    \model{} utilizes the output of the model on the unlabeled target graph to optimize two objectives: Information Maximization and Structure Consistency to adapt the model on the target domain. 
    }
    \label{fig:framework}
\end{figure}
\begin{defn}[\textbf{\emph{Unsupervised Graph Domain Adaptation}}]
\label{def:UGDA}
aims to learn a node classification model $\mathbf{\mathcal{Q}}(\mathbf{y}|\mathbf{G}; \Theta)$ based on a source graph $\mathbf{G_s} = (\mathbf{V_s}, \mathbf{E_s}, \mathbf{X_s}, \mathbf{y_s})$, and the model performs well on a target graph $\mathbf{G_t} = (\mathbf{V_t}, \mathbf{E_t}, \mathbf{X_t}, \mathbf{y_t})$ where node features 
(i.e., $\mathbf{X_s}$ and $\mathbf{X_t}$) and labels (i.e., $\mathbf{y_s}$ and $\mathbf{y_t}$) express the same meanings 
\textbf{$\mathbf{y_t}$ is fully unknown which indicates an unsupervised problem in the target graph.} The oracle conditional distributions for the source and target graph are defined as $p_s(y|x, \mathcal{N})$ and $p_t(y|x, \mathcal{N})$, respectively. 
\end{defn}

A general assumption in Unsupervised Graph Domain Adaptation is that prediction tasks are almost the same, i.e., $p_s(y|x, \mathcal{N})$ and $p_t(y|x, \mathcal{N})$ are similar \cite{zhang2019dane, wu2020unsupervised}. 
Thus, according to Eq.~\eqref{equ:CE}, the main challenges for UGDA are the misalignment between prior distributions $p_s(v)$ of the source graph and $p_t(v)$ of the target graph. Consequently, the majority of current approaches attempt to align two distributions as part of their methodologies, which predominantly depends on access to the source data. Nonetheless, in practical situations, obtaining the training data of the source model is frequently challenging, primarily due to privacy concerns or data collection expenses. This gives rise to a novel problem named \textbf{Source-Free Unsupervised Graph Domain Adaptation (SFUGDA)}, which necessitates the absence of source data access in addition to the requirements of UGDA.

\section{Problem Statement \& Methodology}

\subsection{Problem Statement \& Overview}
\textbf{Problem Statement:} Source Free Unsupervised Graph Domain Adaptation aims to learn a well-performed node classification model 
$\mathbf{\mathcal{Q}_t}(\mathbf{y}|\mathbf{G}; \Theta_t)$ on the target graph $\mathbf{G_t}$, while the accessible information only contains two parts: (1) the well-trained source model $\mathbf{\mathcal{Q}_s}(\mathbf{y}|\mathbf{G}; \Theta_s)$ (i.e., well-performed in the source graph but not guaranteed to be well-performed in the target graph); 
(2) The unlabeled target graph $\mathbf{G_t}=(\mathbf{V}_t, \mathbf{E}_t, \mathbf{X}_t)$. 

\noindent{\textbf{Overview of Framework:}} As the framework outlines shown in Fig. \ref{fig:framework}, the well-performed source model is provided by the first procedure, the inaccessible training procedure on the source graph. 
As we cannot interfere with the source training procedure, \textbf{the source model architecture could be an arbitrary GNN} as we could not determine. 
In our experiments, the well-trained source model is the model with the best performance on the validation set of the source graph. 
Parameters of the source model $\Theta_s$ with primary discriminative ability are utilized as the initialization of our model in the latter adaptation procedure. 

The adaptation procedure is the key to solving this problem. 
In this procedure, we need to further adapt $\mathbf{\mathcal{Q}}_s(\mathbf{y}|\mathbf{G}; \Theta_s)$ on the unlabeled target graph $\mathbf{G}_t$ 
without the information of GNN architecture and the prior distribution of the source data.
This leads to three main challenges: (1) We are required to adapt the source model to the target distribution with no access to features in the source domain; (2) The adaption learning in the target domain is an entirely unsupervised learning procedure because of no access to the label in both source and target domains. 
With only optimizing the unsupervised loss in the target domain, it could be easy to lose the initial discriminatory power of the source model; (3) Algorithm design that depends on the model structure will no longer be feasible because of no access to the source training procedure.
In our work, we design unsupervised optimization objectives to solve the above challenges and achieve better performance on the target graph.

\subsection{Overall Objectives}
To adapt the given source model, we mainly design two optimization objectives. One is to leverage the information stored in the given model, namely the Information Maximization (IM) optimization objective, and the other is to utilize the target graph structure, namely the Structure Consistency (SC) optimization objective, to enhance the discriminative ability of the model on the target graph. The overall objective is defined as follows:
\begin{equation}
\label{equ:overall_objective}
    \max \mathcal{L} = \mathcal{L}_{\im} + \mathcal{L}_{\scim}
\end{equation}
Note that, both optimization objectives are designed on the model output space $\mathbb{R}^k$, where $k$ is the number of classes. The objective can be easily applied to any GNN model. Details on the two objectives are presented in the following sections.

\subsection{Information Maximization Optimization Objective}
We define the IM objective as the mutual information between inputs and outputs of the model enhancing the discriminative ability:
\begin{equation}
    \mathcal{L}_{\im} = \operatorname{MI}(\mathbf{\hat{y}_t}, \mathbf{V_t}) = - \operatorname{H}(\mathbf{\hat{y}_t} | \mathbf{V_t}) + \operatorname{H}(\mathbf{\hat{y}_t}),
\end{equation}
where $\mathbf{\hat{y}_t}$ is the prediction on target domain and $\mathbf{V_t}$ is the information of input nodes containing node feature $\mathbf{X_t}$ and information from node neighbor $\mathbf{\mathcal{N}_t}$. $\operatorname{MI}(\cdot, \cdot)$ is the mutual information, and $\operatorname{H}(\cdot)$ and $\operatorname{H}(\cdot | \cdot)$ are entropy and conditional entropy, respectively. The objective can be divided into two parts, one is to minimize the conditional entropy and the other is to maximize the entropy of the marginal distribution of $\mathbf{\hat{y}_t}$. We will introduce the implementation and the idea behind such a design for these two parts, respectively.

\textbf{Conditional Entropy}
The conditional entropy can be implemented by the following equation:
\begin{equation}
\label{equ:conditional entropy}
    H(\mathbf{\hat{y}_t} | \mathbf{V_t})  =  \mathbb{E}_{x_i \sim p_t(x)} \left[- \sum_{y = 1}^{|\mathcal{C}|} q(y|x_i, \mathbf{\mathcal{N}_i}; \Theta) \log q(y | x_i, \mathbf{\mathcal{N}_i}; \Theta)\right]
\end{equation}
which can be easily optimized by sampling nodes from the prior distribution $p_t(x)$ on the target graph. 

Intuitively, the goal of this objective is to enhance the certainty of predictions made on the target graph, which will lead to a substantial improvement in the lower bound of the model's effectiveness.

Theoretically, we present two key lemmas. The first lemma demonstrates the manner in which the objective bolsters the confidence of the model predictions. Meanwhile, the second lemma indicates that the lower bound of the Area Under Curve (AUC) will increase when the objective is applied.
\begin{lem}
	  \label{lemma:convergence}
	  When the source model is optimized by the objective Eq. \eqref{equ:conditional entropy} with a gradient descent optimizer and the capacity of the source model is sufficiently large, for each node $v_i$ on the target graph, the predicted conditional distribution $q(y|x_i, \mathcal{N}_i; \Theta)$ will converge to a vector \textbf{q}, where the value of $\eta$ elements will be $\frac{1}{\eta}$, and the other elements will be $0$. $\eta$ is determined by the number of categories with the maximum probability value predicted by the original source model $q(y|x_i, \mathcal{N}_i; \Theta_s)$. Similarly, the non-zero positions of \textbf{q} are the indices of categories with the maximum probability value.
\end{lem}
In most cases, $\eta$ equals one, and hence the prediction $\textbf{q}$ will be a one-hot encoding vector. The proof will be listed in the Appendix \ref{app:proof1}.
For further verifying the effectiveness of the objective, we theoretically analyze its effect on the Area Under Curve (AUC) metric on a binary classification problem:
\begin{lem}
	  \label{lemma:auc}
	  When the original source model is trained for a binary classification problem with the discriminative ability of $r_p$ and $r_n$ accuracy for positive samples and negative samples on the target graph respectively, the lower bound of AUC can be raised from $r_p \times r_n$ to $\frac{1}{2}(r_p + r_n)$ by using the conditional entropy objective Eq. \eqref{equ:conditional entropy}. 
\end{lem}
The proof will be listed in the Appendix \ref{app:proof2}. Raising the lower bound of AUC from $r_p \times r_n$ to $\frac{1}{2}(r_p + r_n)$ is significant, for instance, if $r_p = r_n = 0.7$, then the absolute improvement will be $0.21$.

\subsubsection{Entropy of Marginal Distribution}
The entropy of marginal distribution $\mathbf{\hat{y}_t}$ can be calculated as:
\begin{equation}
\label{equ:prediction entropy}
\begin{split}
    H(\mathbf{\hat{y}_t}) =& - \sum_y q(y) \log q(y), \\
    \it{where}\; q(y) =& \; \mathbb{E}_{v_i \sim p_t(v)} \left[q(y|x_i, \mathbf{\mathcal{N}_i}; \Theta) \right].
\end{split}
\end{equation}
This objective is designed to avoid the unsupervised objective easily stuck in a bad solution where predictions concentrate on the same category.
Particularly, if we have additional knowledge about the prior distribution of labels $p_t(y)$ on the target graph, a KL-divergence objective can be a replacement for the Eq. \eqref{equ:prediction entropy}:
\begin{equation}
\label{equ:KL}
    KL(p_t(y)\; ||\; q(y)) = \sum_y p_t(y) \log \frac{p_t(y)}{q(y)}.
\end{equation}

To summarize, we adopt Eq.~\eqref{equ:prediction entropy} by default to balance different categories, and we can use Eq.\eqref{equ:KL} to approximate the real distribution with additional information about the prior label distribution.
 
\subsection{Structure Consistency Optimization Objective}
To adapt the source model to the shifted target domain without source data, leveraging the structural information of the target graph becomes the key solution.  
Thus, we design a Structure Consistency (SC) objective based on two hypotheses, i.e., (1) the probability of sharing the same label for local neighbors is relatively high; (2) the probability of sharing the same label for the nodes with the same structural role is relatively high. These two hypotheses are commonly utilized in lots of Graph Embedding works \cite{perozzi2014deepwalk, grover2016node2vec, ribeiro2017struc2vec} where several structure-preserving losses based on the above hypotheses are designed for learning node representations.
To be specific, the SC objective is designed as follows:
\begin{equation}
\begin{split}
\label{equ:scim}
\mathcal{L}_{SC} = \sum_{v_i \in \mathbf{V_t} }\sum_{v_j \in \mathbf{V_t}}\lambda_1 & \mathcal{J}_{L}^{i,j} + \lambda_2 \mathcal{J}_{C}^{i,j},\\
\mathcal{J}_{L}^{i,j} = p_l^{(i,j)}\log \sigma(\left\langle\mathbf{\hat{y}_t^{(i)}}, \mathbf{\hat{y}_t^{(j)}}\right\rangle) &+ (1-p_l^{(i,j)}) \log \sigma(-\left\langle\mathbf{\hat{y}_t^{(i)}}, \mathbf{\hat{y}_t^{(j)}}\right\rangle),\\
\mathcal{J}_{C}^{i,j} = p_c^{(i,j)}\log \sigma(\left\langle\mathbf{\hat{y}_t^{(i)}}, \mathbf{\hat{y}_t^{(j)}}\right\rangle) &+ (1-p_c^{(i,j)}) \log \sigma(-\left\langle\mathbf{\hat{y}_t^{(i)}}, \mathbf{\hat{y}_t^{(j)}}\right\rangle),
\end{split}
\end{equation}
where $\hat{y}_t^{(i)}$ is the predicted label vector for $i$-th node, $\sigma(x) = 1 / (1 + e^{-x})$ is Sigmoid function, $\left\langle\cdot \;, \cdot \right\rangle$ is the inner product, $\lambda_1$ and $\lambda_2$ are hyperparameters to control the importance of two sub-objectives. Notice that both $\lambda_1$ and $\lambda_2$ are set to the default value 1 in the experiments. $p_l^{(i, j)}, p_s^{(i, j)} \in \{0, 1\}$ are defined by the local neighbor similarity and structural role similarity.
Specifically, if $(v_i, v_j) \in \mathbf{E_t}$, then $p_l^{(i, j)} = 1$, otherwise $p_l^{(i, j)} = 0$. Similarly, if $(v_i, v_j) \in \mathbf{\mathcal{S}_t}$ then $p_c^{(i, j)} = 1$, otherwise $p_c^{(i, j)} = 0$, where $\mathbf{\mathcal{S}_t}$ is a set containing the top $\kappa$ structurally similar node pairs. We follow struc2vec \cite{ribeiro2017struc2vec} to define the structural similarity that can be roughly understood as calculating the similarity of the sorted degree sequences around two given nodes. In order to reduce the number of hyperparameters, $\kappa$ is set as the same size of the edge set $|\mathbf{E_t}|$ by default in all of our experiments while it can be adjusted as needed. 

Intuitively speaking, the first and the second sub-objectives correspond to hypotheses (1) and (2), respectively.$\mathcal{J}^{i,j}_L$ and $\mathcal{J}^{i,j}_C$ are cross-entropy loss defined in the node pair level. The objective $\mathcal{J}_L^{i,j}$ enlarges the prediction similarity between the nodes with connection and distinguishing nodes without connection. Similarly, the objective
$\mathcal{J}_C^{i,j}$ enlarges the similarity between the nodes with similar structural roles and distinguishing nodes with different structural roles.
Finally, we use the negative sampling technique \cite{perozzi2014deepwalk} to avoid calculating the objective function for each node pair for acceleration. 
Combining the objectives Eq. \eqref{equ:overall_objective}, Eq. \eqref{equ:conditional entropy}, and Eq. \eqref{equ:prediction entropy}, we obtain the overall differentiable objective of the model parameters $\Theta$. We adopt the adaptive moment estimation method (i.e., Adam)~\cite{kingma2014adam} to optimize the overall objective.

%% file: src/experiment.tex
\section{Experiments}
We conduct experiments on real-world datasets to study our proposed algorithm \model{}. We design a series of experiments to answer the following research questions:
\begin{itemize}[leftmargin=0.2in]
    \item \textbf{RQ1}: How does the GCN-\model{} compare with other state-of-the-art node classification methods?  
    (GCN-\model{} represents \model{} applying on the default source domain model: GCN.) 
    \item \textbf{RQ2}: 
    How effective can \model{} be integrated with different GNN models?
    \item \textbf{RQ3}: How do different components in \model{} contribute to its effectiveness?
    \item \textbf{RQ4}: How do different choices of hyperparameters $\lambda_1$ and $\lambda_2$ affect the performance of \model{}?
    \item \textbf{RQ5}: Can GCN-\model{} learn more distinguishable node representations from visualization compared with other baselines? 
\end{itemize}

\subsection{Experiment Settings}

\begin{table*}[]
\centering
\caption{\label{tab:result} Average Macro-F1 and Macro-AUC scores on target graph in four unsupervised graph domain adaptation tasks on baseline methods and GCN-\model{}. GCN-\model{}-prior is the GCN-\model{} with additional label distribution knowledge. Notice that, For UGDA baselines including DANE and UDAGCN, we give them additional access to source data.}
\scalebox{0.88}{
\begin{tabular}{c|cc|cc|cc|cc}
\toprule
\multicolumn{1}{c|}{ \multirow{3}*{Methods} }& \multicolumn{4}{c|}{Group1} & \multicolumn{4}{c}{Group2}\\
\cline{2-9}
\multicolumn{1}{c|}{} &\multicolumn{2}{c|}{DBLP$\rightarrow$ACM}&\multicolumn{2}{c|}{ACM$\rightarrow$DBLP}&\multicolumn{2}{c|}{ACM-D$\rightarrow$ACM-S}&\multicolumn{2}{c}{ACM-S$\rightarrow$ACM-D}\\
\cline{2-9}
\multicolumn{1}{c|}{} & Macro-F1 & Macro-AUC & Macro-F1 & Macro-AUC & Macro-F1 & Macro-AUC & Macro-F1 & Macro-AUC \\
\hline
DeepWalk & 0.135 $\pm$ 0.012 & 0.593 $\pm$ 0.010 & 0.112 $\pm$ 0.012 & 0.613 $\pm$ 0.008 & 0.183 $\pm$ 0.012 & 0.549  $\pm$ 0.006 & 0.237 $\pm$ 0.012 & 0.573  $\pm$ 0.007\\
Node2vec & 0.128 $\pm$ 0.023 & 0.567 $\pm$ 0.011 & 0.080 $\pm$ 0.018 & 0.533 $\pm$ 0.004 & 0.134 $\pm$ 0.012 & 0.537  $\pm$ 0.005 & 0.219 $\pm$ 0.014 & 0.649  $\pm$ 0.003\\
\hline
GCN  & 0.583 $\pm$ 0.002 & 0.887 $\pm$ 0.004 & 0.668 $\pm$ 0.015 & 0.937 $\pm$ 0.003 & 0.685 $\pm$ 0.005 & 0.856 $\pm$ 0.008 & 0.796 $\pm$ 0.030 & 0.924 $\pm$ 0.002\\
GraphSAGE & 0.418 $\pm$ 0.057 & 0.763 $\pm$ 0.054 & 0.752 $\pm$ 0.010 & 0.934 $\pm$ 0.003 & 0.407 $\pm$ 0.042 & 0.835 $\pm$ 0.013 & 0.743 $\pm$ 0.015 & 0.909 $\pm$ 0.003 \\   
GAT & 0.227 $\pm$ 0.004 & 0.831 $\pm$ 0.004  & 0.745 $\pm$ 0.036 & 0.929 $\pm$ 0.011 &  0.681 $\pm$ 0.006 & 0.854 $\pm$ 0.005 & 0.804 $\pm$ 0.007 & 0.928 $\pm$ 0.002 \\

\hline
GRACE & 0.604 $\pm$ 0.014 & 0.908 $\pm$ 0.007 & 0.604 $\pm$ 0.014 & 0.806 $\pm$ 0.004 & 0.662 $\pm$ 0.003 
& 0.876 $\pm$ 0.003 & 0.792 $\pm$ 0.007 & 0.907 $\pm$ 0.007 \\
DGI  & 0.592 $\pm$ 0.010 & 0.894 $\pm$ 0.012 & 0.621 $\pm$ 0.005 & 0.872 $\pm$ 0.010 & 0.610 $\pm$ 0.003 & 0.842 $\pm$ 0.002 & 0.808 $\pm$ 0.006 & 0.919 $\pm$ 0.007\\
TENT & 0.617 $\pm$ 0.007 & 0.912 $\pm$ 0.008 & 0.913 $\pm$ 0.011 & 0.957 $\pm$ 0.009 & 0.702 $\pm$ 0.015 & 0.893 $\pm$ 0.004 & 0.813 $\pm$ 0.007 & 0.922 $\pm$ 0.004 \\
GTrans & 0.610 $\pm$ 0.003 & 0.913 $\pm$ 0.006 & 0.911 $\pm$ 0.007 & 0.948 $\pm$ 0.004 & 0.723 $\pm$ 0.021 & 0.864 $\pm$ 0.003 & 0.753 $\pm$ 0.004 & 0.876 $\pm$ 0.018 \\

\hline
SHOT & 0.556 $\pm$ 0.004 & 0.858 $\pm$ 0.007 & 0.673 $\pm$ 0.079 & 0.938 $\pm$ 0.008 & 0.658 $\pm$ 0.011 & 0.864 $\pm$ 0.002 & 0.827 $\pm$ 0.016 & 0.916 $\pm$ 0.018 \\
NRC  & 0.561 $\pm$ 0.009 & 0.858 $\pm$ 0.009 & 0.644 $\pm$ 0.010 & 0.897 $\pm$ 0.007 & 0.629 $\pm$ 0.007 & 0.823 $\pm$ 0.010 & 0.817 $\pm$ 0.003 & 0.921 $\pm$ 0.005\\ \hline

\hline
DANE & 0.614 $\pm$ 0.017 & 0.906 $\pm$ 0.023 & 0.584 $\pm$ 0.008 & 0.937 $\pm$ 0.003 & 0.722 $\pm$ 0.004 & 0.888 $\pm$ 0.002 & 0.821 $\pm$ 0.004 & 0.923 $\pm$ 0.001 \\
UDAGCN  & 0.626 $\pm$ 0.070 & 0.930 $\pm$ 0.006 & 0.696 $\pm$ 0.009 & 0.953 $\pm$ 0.008 & 0.665 $\pm$ 0.010 & 0.881 $\pm$ 0.004 & 0.822 $\pm$ 0.018 & 0.928 $\pm$ 0.002\\ \hline
\textbf{GCN-\model{}}  & \underline{0.636 $\pm$ 0.003} & \underline{0.931 $\pm$ 0.004}&\underline{ 0.928 $\pm$ 0.018} & \underline{0.988 $\pm$ 0.002} & \underline{0.733 $\pm$ 0.005} & \underline{0.907 $\pm$ 0.005} & \underline{0.842 $\pm$ 0.008} & \underline{0.951 $\pm$ 0.002} \\ 
GCN-\model{}-prior  & \textbf{0.650 $\pm$ 0.007} & \textbf{0.943 $\pm$ 0.008}& \textbf{0.935 $\pm$ 0.011} &\textbf{ 0.990 $\pm$ 0.001} & \textbf{0.737 $\pm$ 0.004} & \textbf{0.908 $\pm$ 0.005} & \textbf{0.843 $\pm$ 0.003} & \textbf{0.953 $\pm$ 0.005} \\\bottomrule
\end{tabular}}
\end{table*}

\paragraph{Datasets.}
We use two groups of real-world graph datasets for our experiments.
DBLPv8 and ACMv9 are the first group of citation networks collected by \cite{wu2020unsupervised} from arnetMiner \cite{tang2008arnetminer}. 
Their domain gap mainly comes from different origins (DBLP, ACM respectively) and different publication time periods, i.e. DBLPv8 (after 2010), ACMv9 (between years 2000 and 2010).
ACM-D (Dense) and ACM-S (Sparse) are the second group of citation networks from the ACM dataset collected by \cite{yang2020unsupervised}.
The detailed statistics of these datasets are illustrated in Tab. \ref{tab:datasets}. More details can be found in Appendix~\ref{app:dataset}.

\begin{table}[tbp]
    \centering
    \caption{\label{tab:datasets}Statistics of the experimental datasets}
    \begin{tabular}{c|cccc}
        \hline
        Datasets & \# Nodes & \# Edges & \# Features & \# Labels \\ \hline
        DBLPv8 & 5578 & 7341 & 7537 & 6\\
        ACMv9 & 7410 & 11135 & 7537 & 6\\ \hline
        ACM-D & 1500 & 4960 & 300 & 4 \\
        ACM-S & 1500 & 759 & 300 & 4 \\ \hline
    \end{tabular}
\end{table}

\paragraph{Baselines.}
We select some state-of-the-art methods as baselines to verify the effectiveness of our proposed algorithm. 
They are (1) Graph embedding methods including DeepWalk, and Node2vec \cite{perozzi2014deepwalk,  grover2016node2vec}. (2) Graph Neural Network (GNN) methods including GCN, GraphSAGE, and GAT \cite{kipf2016semi, hamilton2017inductive, velivckovic2017graph}. 
(3) SFUDA methods on the image including SHOT~\cite{liang2020we} and NRC~\cite{yang2021exploiting} 
(4) self-supervised learning methods including DGI~\cite{velickovic2019deep}, GRACE~\cite{zhu2020deep}, TenT~\cite{wang2020tent}, and Gtrans~\cite{jin2022empowering}. DGI and GRACE are two graph self-supervised learning baselines that focus on learning good representation for the downstream task. However, they require labeled data on the target domain for finetuning, which is not available in our scenario. To make a fair comparison, we employ the proposed self-supervised learning algorithm for unsupervised fine-tuning, which shows a similar train manner with our proposed \model{}.
TenT and Gtrans are the selected self-supervised learning algorithms for unsupervised finetuning from image and graph domains, respectively. 
(5) UGDA methods including DANE and UDAGCN \cite{zhang2019dane, wu2020unsupervised}. \textbf{Notice that we give them UGDA methods additional access to the source data.}
For all baseline methods with GNN, we utilize a two-layer GCN in which the hidden dimensions are 256, and 128 respectively as the default encoder.
All baseline methods are included with a careful hyperparameter check. Details can be found in the Appendix~\ref{app:experiment}

\paragraph{Reproducibility Settings.}
To ensure the validity of experiments, each source dataset is randomly split into training and validation sets with a ratio 4:1.
All the experimental results are averaged over 5 runs with different random seeds [1, 3, 5, 7, 9]. 

Considering the hyperparameter search setting, \textbf{we do not have any hyperparameter search in \model{}}. We set the hyperparameters $\lambda_1$ and $\lambda_2$ to the default value: of 1 in all experiments except for the hyperparameter sensitivity analysis.
For baseline methods, we apply large-scale hyperparameter grid search on all other baselines including GNN, SFUDA, self-supervised learning, and UGDA methods to ensure those baseline methods reach their best performance. 
Our reproducing settings are different from some baseline methods reported in their papers like five different random seeds and the validation partition.
After checking with their authors, we consider that this may induce a different experimental result from the one in the original paper.
The details of experimental settings can be found in the appendix \ref{app:experiment}.
We also release our code and the experimental details in the repository \href{https://github.com/HaitaoMao/SOGA}{here}

\paragraph{Evaluation Methods.}
We conduct the stability evaluation to verify the stability of algorithms. 
The main reason is that models trained with different epochs may have large performance differences. Generally speaking, there should be an additional validation set used to select the best training epoch. 
However, the validation set is not available on the unlabeled target domain. 
Therefore, it is of great importance to evaluate the stability of model performance on the target domain. 
With good stability, it is easy for models to achieve satisfying performance for stability prevents significant performance fluctuations after convergence.
Contrastively, the performance of the unstable method will drop quickly after reaching the peak or fluctuate continuously.
Concretely speaking, we (1) plot the line chart describing the Macro-F1 score on the target domain in each training epoch. 
(2) calculate the mean and standard deviation of Macro-F1 scores on the target domain across the epochs. 
To avoid the initial fluctuation before convergence, we choose the epochs after the first $n$ ones for calculation (set to 20 by default).
A flat curve with little fluctuation indicates good stability, corresponding to results with large expectations and small standard deviations.

\begin{table*}[tbp]
\centering
\caption{\label{tab:result_agnostic} The performance comparison on the Macro-F1 score of different GNN models with or without applying \model{}.}
\begin{tabular}{c|cc|cc}
\toprule
\multicolumn{1}{c|}{ \multirow{2}*{Methods} }& \multicolumn{2}{c|}{Group1} & \multicolumn{2}{c}{Group2}\\
\cline{2-5}
& DBLPv8$\to$ACMv9 & ACMv9$\to$DBLPv8 & ACM-D$\to$ACM-S & ACM-S$\to$ACM-D \\
\hline
GCN  & 0.583 $\pm$ 0.002 & 0.668 $\pm$ 0.015 & 0.685 $\pm$ 0.005 & 0.796 $\pm$ 0.030 \\
\textbf{GCN-\model{}}  & \textbf{0.636 $\pm$ 0.003} & \textbf{0.928 $\pm$ 0.018} & \textbf{0.736 $\pm$ 0.007} & \textbf{0.838 $\pm$ 0.008} \\\hline
GraphSAGE & 0.418 $\pm$ 0.057 & 0.752 $\pm$ 0.010 & 0.407 $\pm$ 0.042 & 0.743 $\pm$ 0.015 \\   
\textbf{GraphSAGE-\model{}}  & \textbf{0.594 $\pm$ 0.086} & \textbf{0.947 $\pm$ 0.002} & \textbf{0.734 $\pm$ 0.006} & \textbf{0.820 $\pm$ 0.020} \\ \hline
GAT & 0.227 $\pm$ 0.004 & 0.745 $\pm$ 0.036 & 0.681 $\pm$ 0.006 & 0.804 $\pm$ 0.007 \\
\textbf{GAT-\model{}}  & \textbf{0.592 $\pm$ 0.086} & \textbf{0.946 $\pm$ 0.001} & \textbf{0.736 $\pm$ 0.006} & \textbf{0.824 $\pm$ 0.027} \\\bottomrule
\end{tabular}
\end{table*}

\subsection{Overall Results (RQ1).} 
The experimental results of all baseline methods, GCN-\model{} (applying \model{} on GCN), and GCN-\model{}-prior on Macro-F1 score and Macro-AUC score are illustrated on Tab. \ref{tab:result}. 
\begin{figure}[!h]
    \centering
    \includegraphics[width=0.4\textwidth]{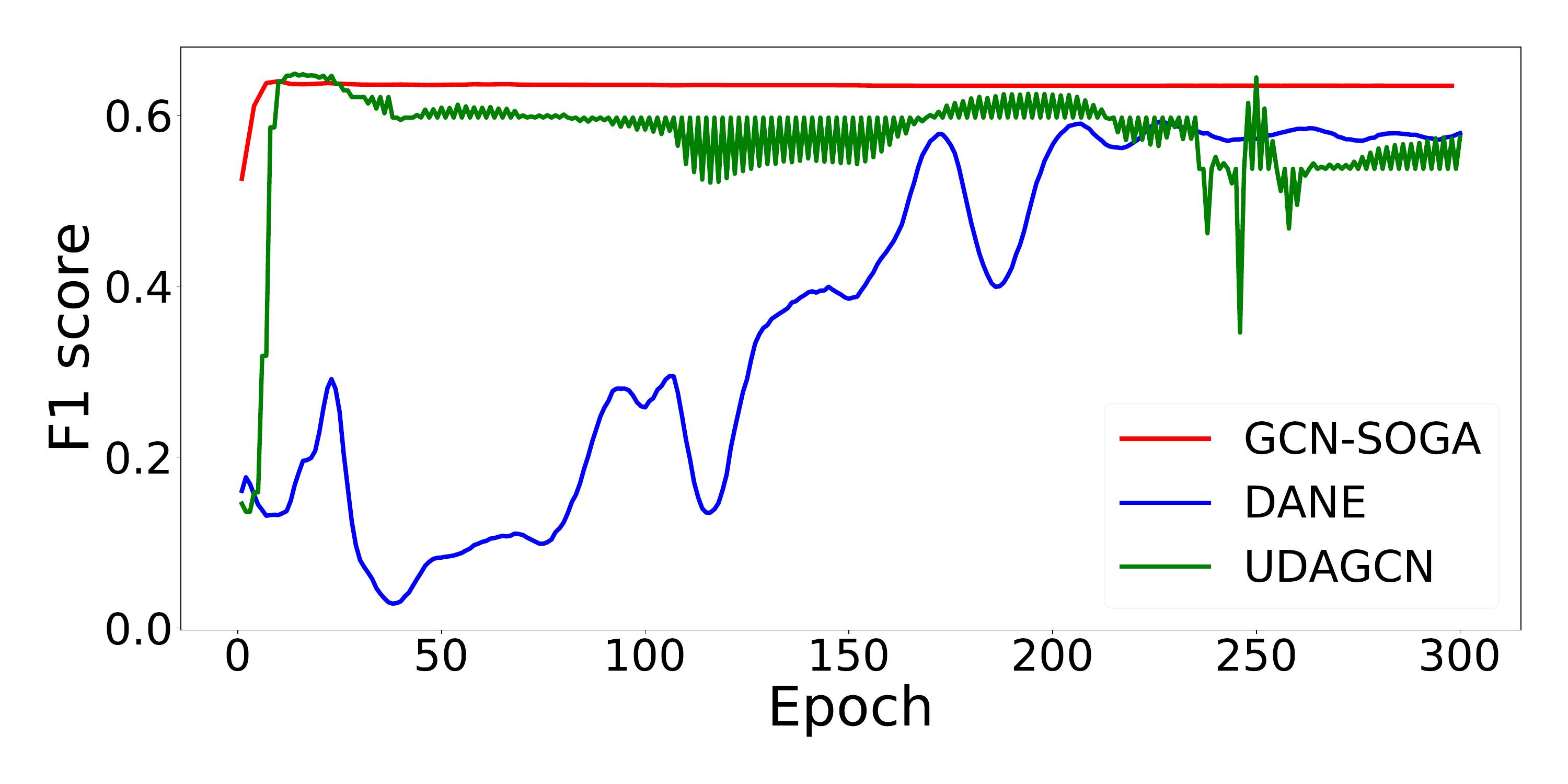}
    \centering
    \caption{
   The comparison of learning curves of different UGDA methods on the DBLPv8 $\to$ ACMv9 task. The $x$ axis denotes the training epoch, whereas the $y$ axis denotes the Macro-F1 score on the target graph.}
    \label{fig:baseline}
\end{figure}

GCN-\model{}-prior is a variant of GCN-\model{} which we give \model{} algorithm with additional knowledge about the prior distribution of labels $p_t(y)$. Therefore, we replace the entropy of marginal distribution in eq. \eqref{equ:prediction entropy} by default with the KL-divergence objective in eq. \eqref{equ:prediction entropy}.
Notice that UGDA methods other than our proposed GCN-\model{} require additional information, i.e., access permission to the source graph in the adaptation procedure. 
Therefore, these methods are not feasible in the SFUGDA scenario. When reproducing these methods,\textbf{ we give them additional access to the source data.}

Experimental results of GCN-\model{} on four cross-domain tasks show consistent improvements in Macro-F1 score and Macro-AUC score, with a maximum gain of 2.1\% and 4\%, respectively. Additionally, GCN-\model{}-prior can have greater gain than GCN-\model{} on the first group.  It indicates that SOGA can work better with prior label distribution awareness. The reason for the performance GCN-\model{}-prior is similar to GCN-\model{} is that both ACM-D and ACM-S datasets have an almost uniform label distribution $p(y)$. The additional prior knowledge happens to be close to the default assumption. Therefore, it is no wonder that the performance of GCN-\model-prior is similar to the one of GCN-\model{}.
From the perspective of different cross-domain tasks, we can find the performance on the DBLPv8 $\to$ ACMv9 and ACM-D $\to$ ACM-S is much lower than the other two tasks which indicates its difficulties. We will mainly focus on these difficult tasks and conduct further experiments on them in later sections.

\begin{figure}[ht]
    \centering
    \includegraphics[width=0.4\textwidth]{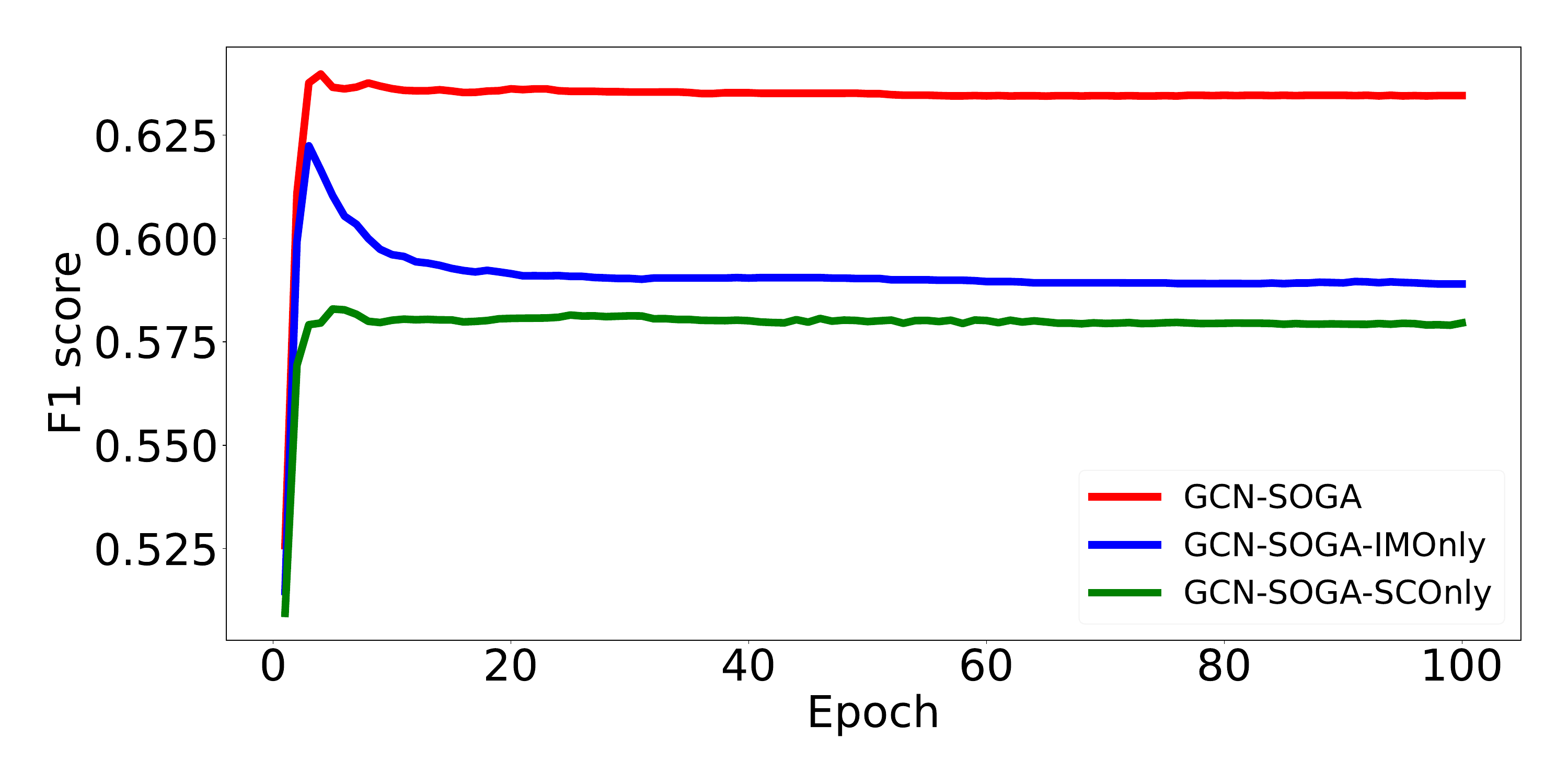}
    \caption{\label{fig:ablation} The comparison of learning curves of GCN-\model{} and its variants on the DBLPv8 $\to$ ACMv9 task. The $x$ axis denotes the training epoch, whereas the $y$ axis denotes the Macro-F1 score in the target graph.}
\end{figure}
From the perspective of different baseline methods, we can see that the graph embedding methods perform poorly, probably due to the lack of preserving cross-graph similarity. 
Though the relative position between nodes is preserved by structural proximity, similar nodes may have entirely different absolute positions in different graphs.
GNNs perform better for the message passing mechanism like graph convolution and can preserve the similarity of nodes if their local sub-graphs are similar as proven by \cite{donnat2018learning}. 
Self-supervised learning methods DGI and GRACE do not perform well since they are designed to learn a good representation but not the discriminative ability on the specific downstream task. 
TENT also does not work so well since it does not take the complex relationship on the graph into consideration. 
GTrans shows unsatisfying performance on ACM-D$\to$ACM-S task since it transforms the target graph structure into a more sparse one. However, this sparsity assumption does not always come true in reality.
SFUDA methods on the image domain including SHOT and NRC do not show satisfying performance since they do not have the specific design on the graph.
Two UGDA methods, UDAGCN and DANE, perform best among all baselines for they implicitly mitigate the distribution gap.
To give a more careful comparison between UGDA baselines and our proposed GCN-\model{}, we conduct stability evaluation in detail.
From Fig. \ref{fig:baseline}, we can see that GCN-\model{} in red illustrates stability with a high Macro-F1 score, while DANE in blue reveals a slower convergence. The performance of UDAGCN in green illustrates a violent fluctuation. The fluctuation majorly contributes to the conflict in optimizing the domain alignment loss and cross-entropy loss. Such fluctuation will cause great difficulty in deciding which epoch to stop the adaptation procedure. 
Thus, the result further indicates the strength of our GCN-\model{}. 
Moreover, such fluctuation indicates that the alignment loss may have a strong conflict with the main cross-entropy loss during the optimization. 
It could be the evidence of why those UGDA methods with additional access to the source data perform worse than our \model{}.

\begin{table*}[ht]
\centering
\caption{\label{tab:ablation study} Statistical results of stability evaluation on GCN-\model{} and its variants. The reported results are the expectation and standard derivation of Macro-F1 scores on the training procedure after 20 epochs}
\begin{tabular}{c|cc|cc}
\toprule
\multicolumn{1}{c|}{ \multirow{2}*{Methods} }& \multicolumn{2}{c|}{Group1} & \multicolumn{2}{c}{Group2}\\
\cline{2-5}

& DBLPv8$\to$ACMv9 & ACMv9$\to$DBLPv8 & ACM-D$\to$ACM-S & ACM-S$\to$ACM-D \\
\hline
GCN  & 0.5832 $\pm$ 0.0000 & 0.6683 $\pm$ 0.0000 & 0.6857 $\pm$ 0.0000 & 0.7961 $\pm$ 0.0000 \\
\hline
GCN-\model{}  & 0.6151 $\pm$ 0.0005 & 0.9382 $\pm$ 0.0002 & 0.7323 $\pm$ 0.0017 & 0.8244 $\pm$ 0.0056 \\
GCN-\model{}-IMOnly  & 0.5823 $\pm$ 0.0007 & 0.9406 $\pm$ 0.3640 & 0.7227 $\pm$ 0.0170 & 0.8263 $\pm$ 0.0060 \\
GCN-\model{}-SCOnly  & 0.5576 $\pm$ 0.0004 & 0.6491 $\pm$ 0.0653 & 0.7160 $\pm$ 0.0026 & 0.3182 $\pm$ 0.0013 \\
\bottomrule
\end{tabular}
\end{table*}

\begin{figure*}[!h]
    \subfigure[The performance on the DBLPv8 $\to$ ACMv9 task]{
    \label{fig:hyper1}
    \centering
    \begin{minipage}[b]{0.45\textwidth}
    \includegraphics[width=1.0\textwidth]{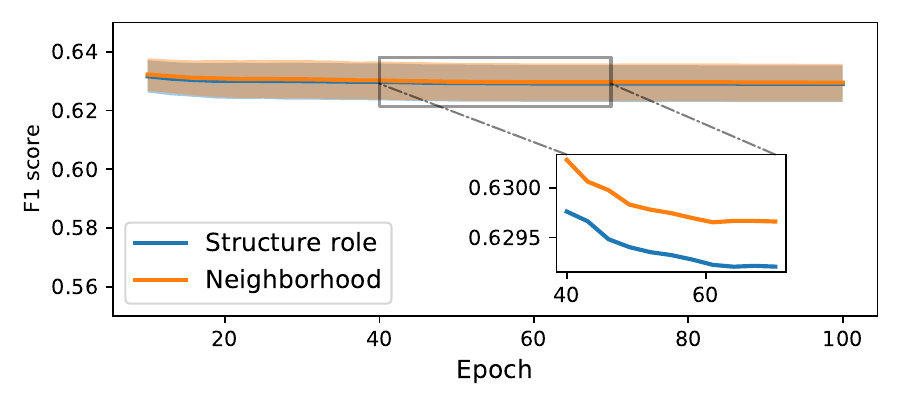}
    \end{minipage}
    }
    \subfigure[The performance on the ACM-D $\to$ ACM-S task]{
        \label{fig:hyper2}
        \centering
        \begin{minipage}[b]{0.45\textwidth}
        \includegraphics[width=1.0\textwidth]{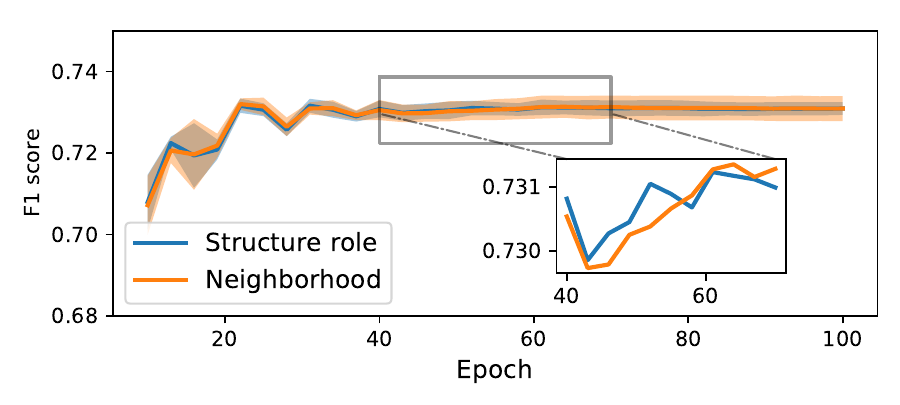}
        \end{minipage}
    }
    \caption{\label{fig:hyper} 
    Performances on the target domain with different choices of $\lambda_1$ and $\lambda_2$. The orange solid line and the corresponding shadow indicate the mean value and the standard deviation of the results, respectively, when $\lambda_2 > \lambda_1$, which means more attention on the neighborhood. 
    The blue ones are similar except $\lambda_1 > \lambda_2$
    }
\end{figure*}

\subsection{Effectiveness of \model{} on different GNN models (RQ2). }
To demonstrate the efficacy and the model agnostic property of our proposed algorithm: \model{}, we evaluate \model{} with different representative GNN models. Specifically, we combine \model{} with GCN, GraphSAGE and GAT, named GCN-\model{}, SAGE-\model{}, and GAT-\model{}, respectively. 
The results are shown in Tab.~\ref{tab:result_agnostic}.
One observation is that \model{} can bring consistent improvement on different GNN models, which verifies that \model{} is model-agnostic. 
Meanwhile, the poor performance of GAT and GraphSAGE on some tasks (e.g., that in ACM-D $\to$ ACM-S) with the careful hyperparameter grid search, indicates the potential overfitting problem of these expressive models. 
Nonetheless, on the poor performance like GraphSAGE on the ACM-D $\to$ ACM-S task, the original Macro-F1 performance is merely 0.407 while performance with \model{} is 0.734, almost the same with the best result: 0.736. 
It indicates that \model{} can help to achieve a comparable good performance regardless of the poor origin GNN model.

\begin{figure*}[!h]
    \subfigure[GCN]{
    \centering
    \begin{minipage}[b]{0.18\textwidth}
    \includegraphics[width=1.0\textwidth]{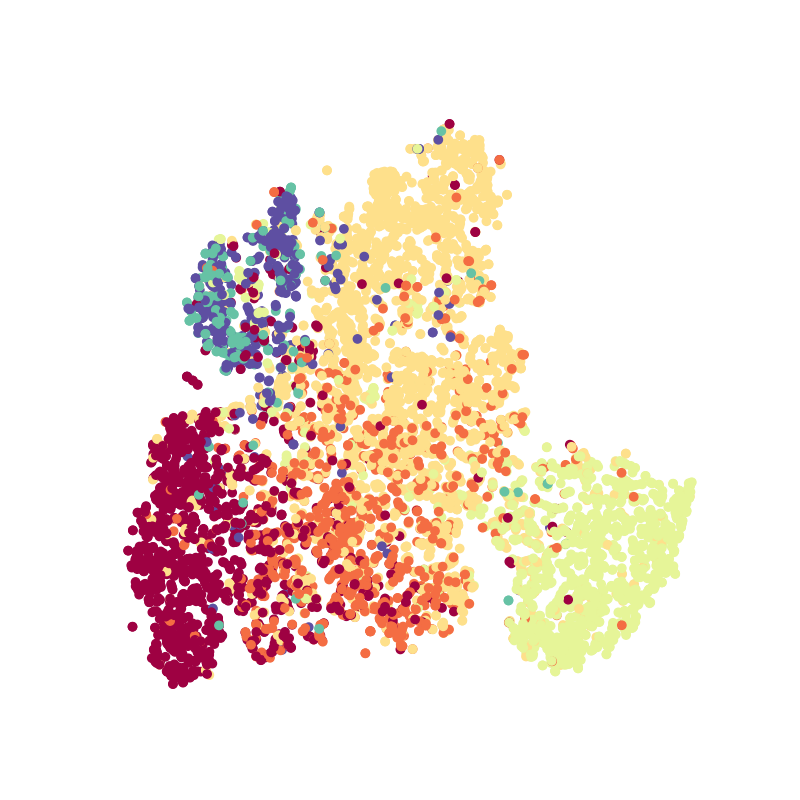}
    \end{minipage}
    }
    \centering
    \subfigure[DANE]{
        \centering
        \begin{minipage}[b]{0.18\textwidth}
        \includegraphics[width=1.0\textwidth]{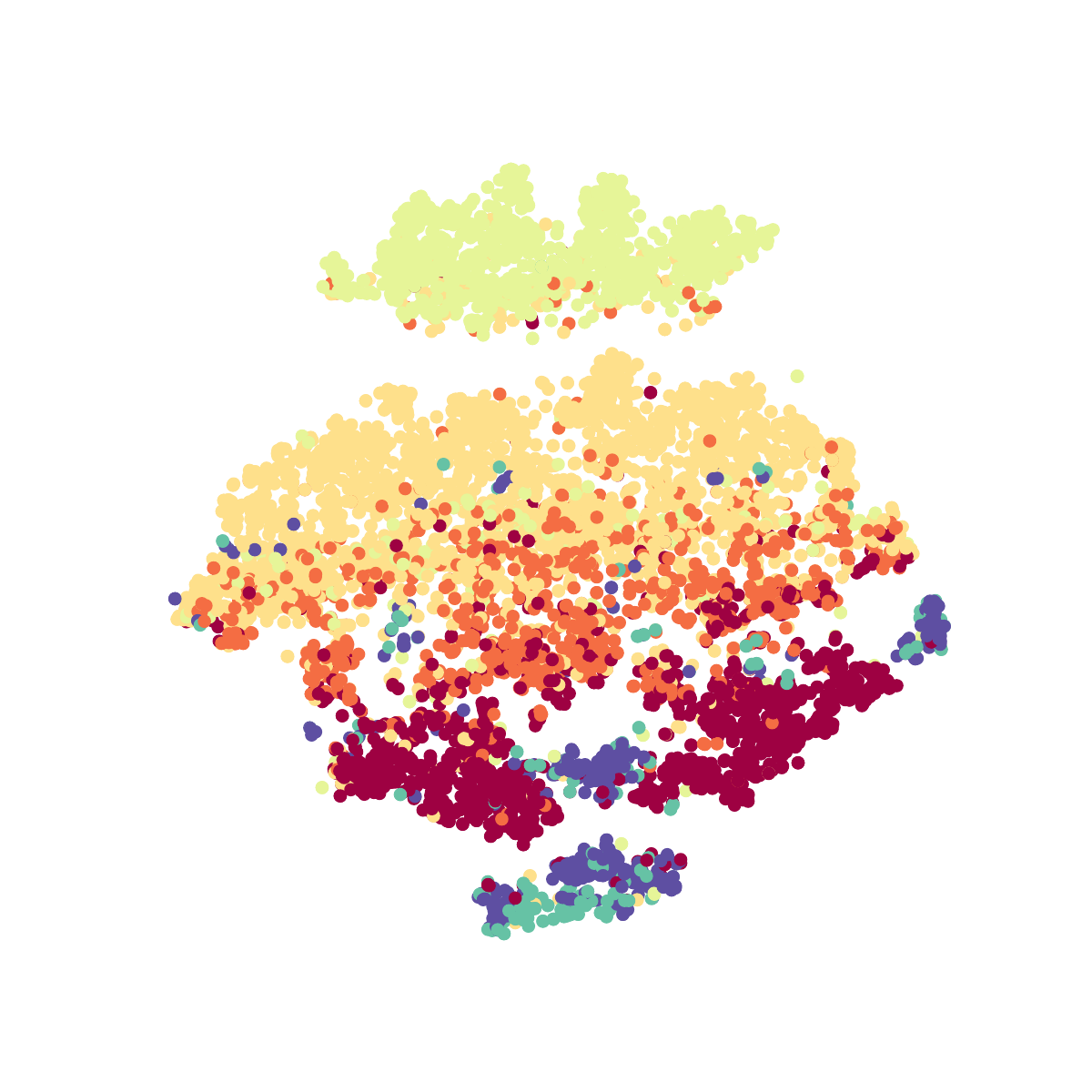}
        \end{minipage}
    }
    \subfigure[UDAGCN]{
        \centering
        \begin{minipage}[b]{0.18\textwidth}
        \includegraphics[width=1.0\textwidth]{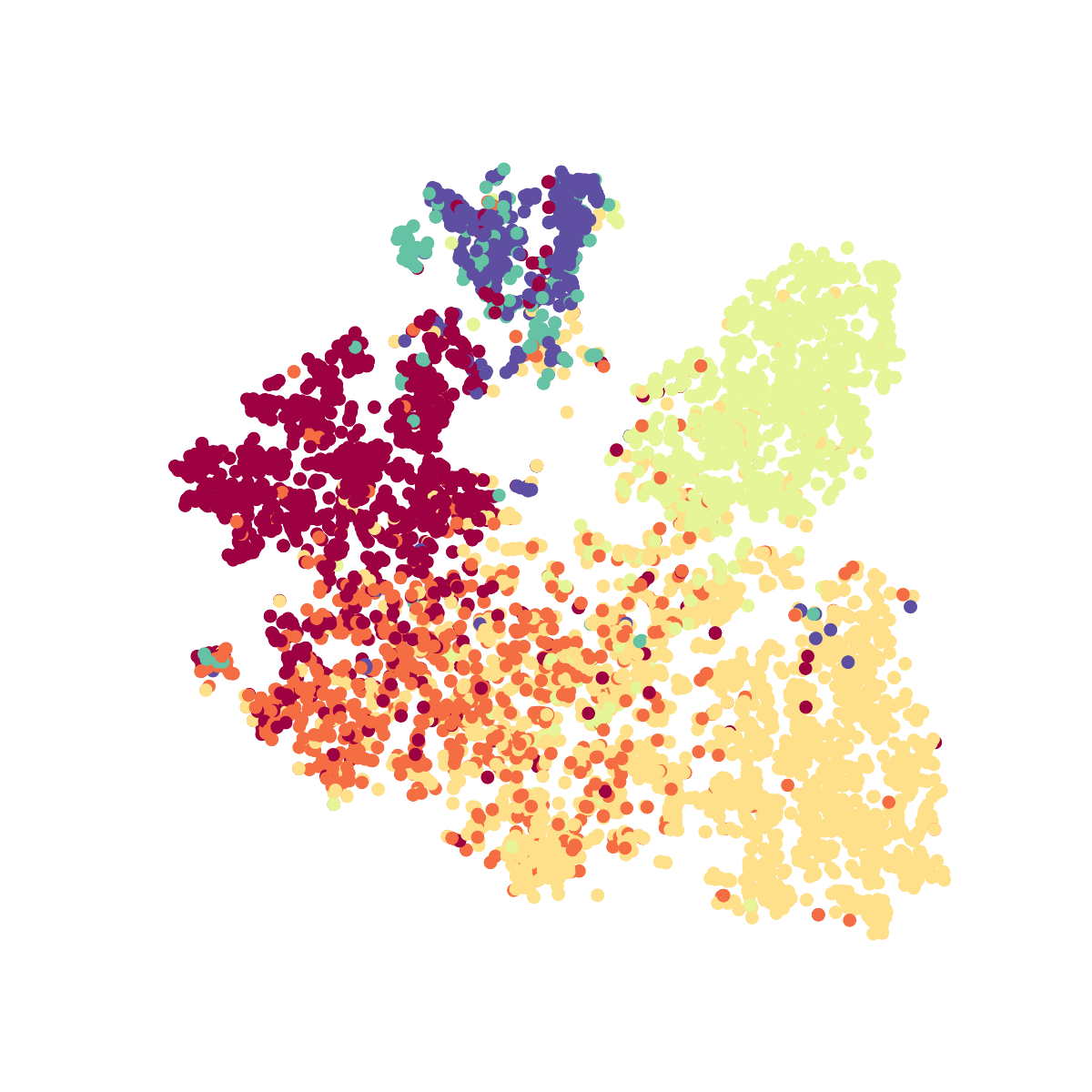}
        \end{minipage}
    }
    \subfigure[GCN-\model{}-IMOnly]{
        \centering
        \begin{minipage}[b]{0.18\textwidth}
        \includegraphics[width=1.0\textwidth]{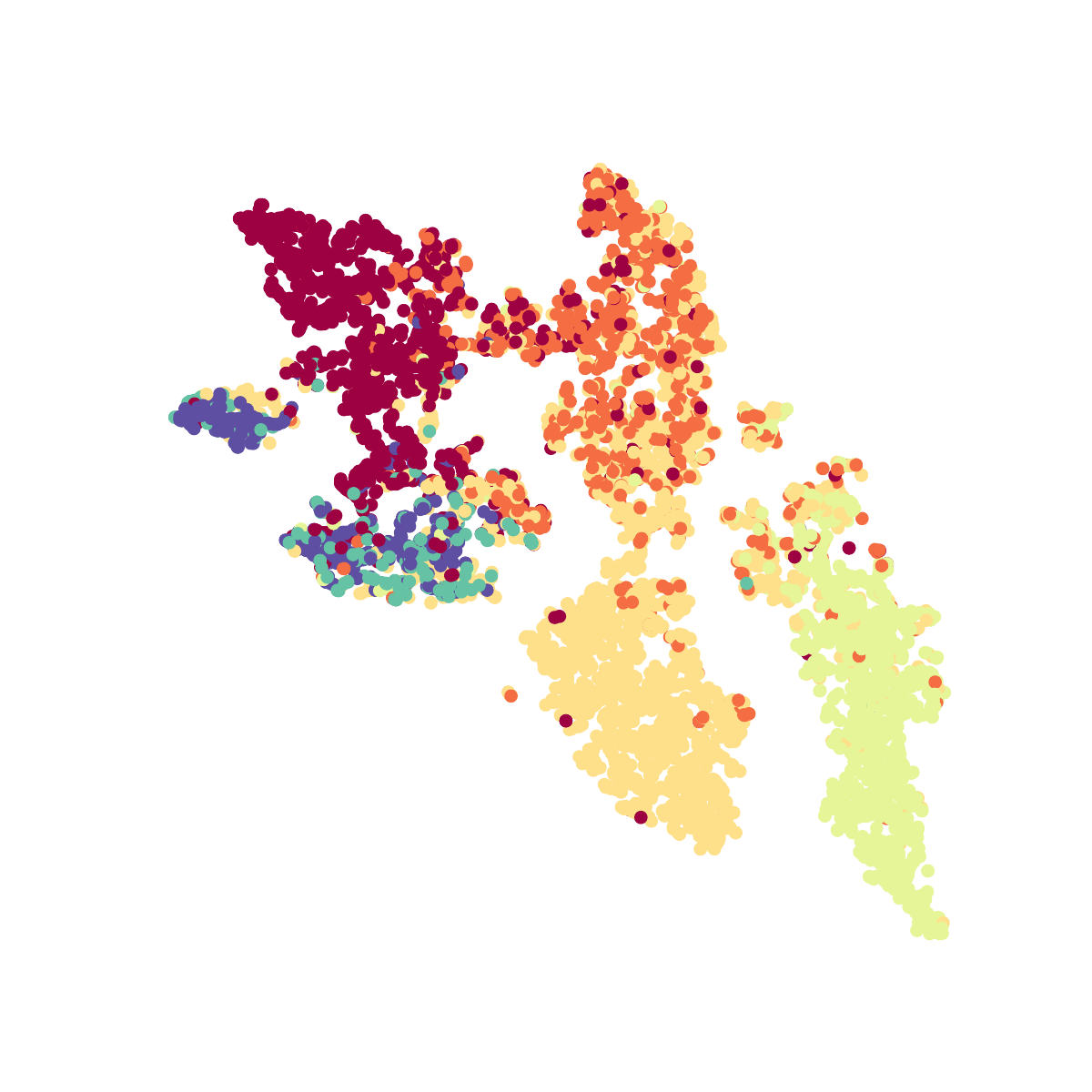}
        \end{minipage}
    }
    \subfigure[\textbf{GCN-\model{}}]{
        \centering
        \begin{minipage}[b]{0.18\textwidth}
        \includegraphics[width=1.0\textwidth]{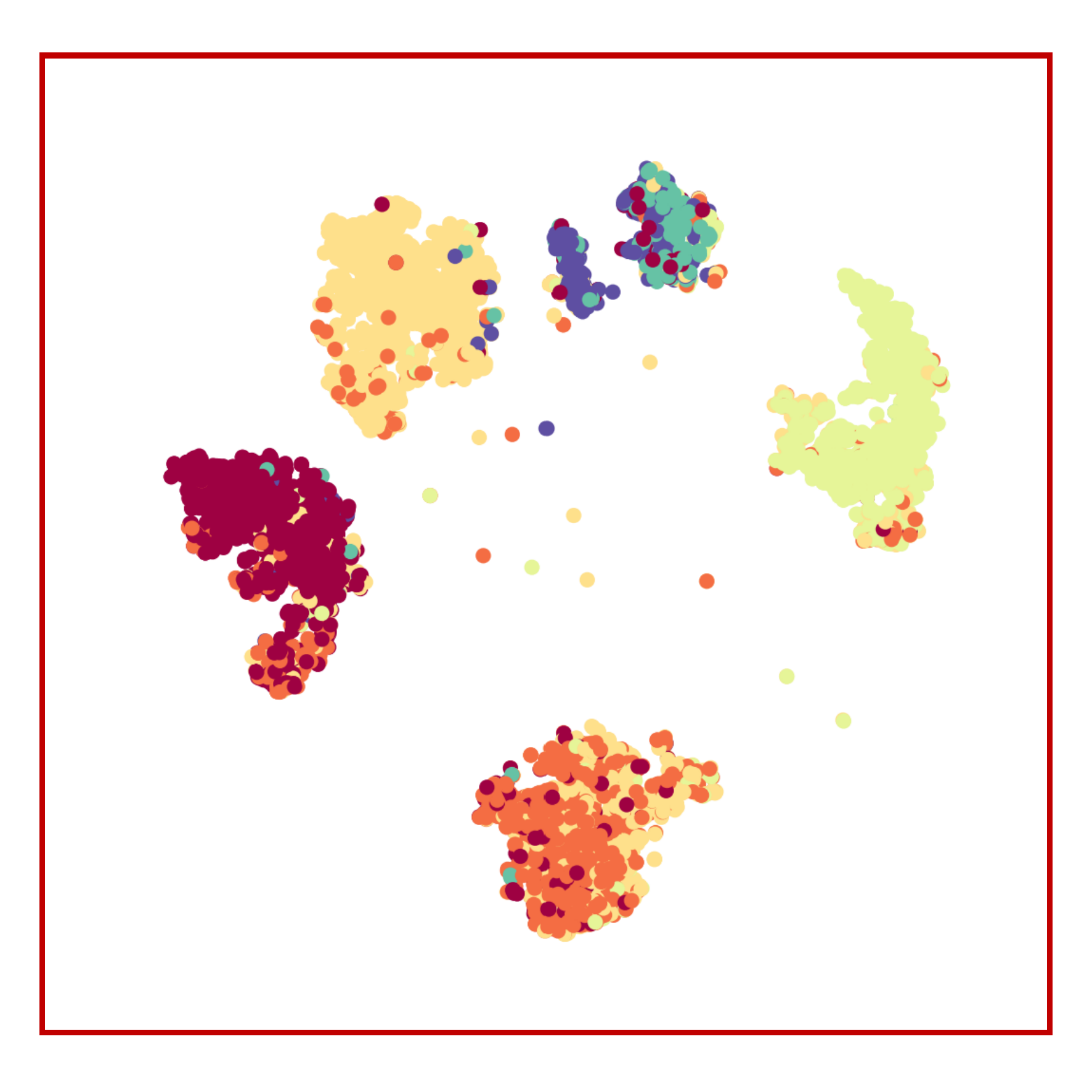}
        \end{minipage}
    }    
    \caption{\label{fig:visual1} Target domain network representation visualization of the source GCN and UGDA methods using t-SNE on the task DBLPv8 $\to$ ACMv9. Our proposed GCN-\model{} is marked with the red box. GCN-\model{}-IMOnly is a variant of GCN-\model{} mentioned in \ref{sec:ablation} without SC optimization objective}
\end{figure*}

\subsection{Ablation Study (RQ3). \label{sec:ablation}}
We conduct ablation experiments to investigate the contribution of each component. The study can be divided into two parts including the necessity of the well-trained source model, and the roles of two optimization objectives.  All experiments in this section utilize the stable evaluation for a more detailed and fair comparison.

First, we verify the necessity of the indispensable component: the well-trained source model. 
Experiments show that only applying two unsupervised objectives with a randomly initialized model leads to failure where the highest Macro-F1 score among all tasks is no more than 0.20, similar to random guess. 
Therefore, it is of great significance to utilize the primary discriminative ability of the well-trained source model. 

Then we further explore the different roles of two unsupervised optimization objectives by stability evaluation. 
We propose the variants of our proposed \model{} called \model{}-IMOnly and \model{}-SCOnly corresponding to the algorithm trained with only IM or SC objective, respectively.  
The curves on DBLPv8 $\to$ ACMv9 are shown in Fig. \ref{fig:ablation}.
We can find the following observations:
(1) For our GCN-SOGA in red, shows significant performance gain and strong stability with a flat curve after the first few epochs. 
(2) For GCN-SOGA-IMOnly in blue, there has been an evident drop after reaching the peak around 10 epochs. This unstable phenomenon reveals the difficulty and uncertainty of achieving a good result. 
(3) For GCN-SOGA-SCOnly in green, the curve is smooth after reaching the peak with only a little gain. 
Overall speaking, IM and SC objectives show a complementary effect. IM enhances the discriminative ability to achieve better results while SC takes charge of maintaining stability to maintain good performance consistently in the training procedure. 

The detailed statistical results of stability evaluation on all tasks are illustrated in Tab. \ref{tab:ablation study}.
We can find that on easy tasks like ACMv9 $\to$ DBLPv8, GCN-SOGA-IMOnly can achieve similar results with GCN-SOGA.
However, on more difficult tasks, which are DBLPv8 $\to$ ACMv9 and ACM-D $\to$ ACM-S, GCN-SOGA could achieve more significant and stable improvement than GCN-SOGA-IMOnly. 
It further indicates the necessity of SC especially on more difficult tasks. 
Moreover, we notice that GCN-SOGA-SCONLY may perform even worse than the original GCN trained on the source domain. 
The key reason is that we could not utilize any supervised signal in the SFUGDA scenario. 
Therefore, both SC and IM objectives are unsupervised objectives. 
It could be possible that we lose the original discriminative ability with unsupervised finetuning.

\subsection{Hyperparameter sensitivity analysis. (RQ4)} 
Though we have achieved impressive results with the default hyperparameter setting as $\lambda_1 = \lambda_2 = 1$, it is still noteworthy to examine how the different choices of $\lambda_1$ and $\lambda_2$ affect the performance of \model{}.
Specifically, we focus on the Macro-F1 score performance of GCN-\model{} as well as its stability in the training procedure with different hyperparameter settings.
Revolving around this goal, we conduct neighborhood evaluation and structure role evaluation which spare larger weight to $\lambda_1$ and $\lambda_2$, respectively.

For neighborhood evaluation, we run experiments 10 times with different choices where $\lambda_1 > \lambda_2$. 
Details of hyperparameter choices are shown in Appendix \ref{app:experiment}. 
Then we plot the line chart describing the Macro-F1 score on the target domain after the first 10 training epochs, skipping initial fluctuations for brevity.
Structure role evaluation is similar except $\lambda_2 > \lambda_1$.
The experimental results on DBLPv8 $\to$ ACMv9 and ACM-D $\to$ ACM-S are illustrated in Fig. \ref{fig:hyper}. 
The solid line is the average result of 10 experiments while the shadow one represents the corresponding standard deviation. 
We can see that curves stay at a high level and the shadow area is narrow, which indicates good performance with stability in the training procedure. We can conclude that our performance is robust to different choices of $\lambda_1$ and $\lambda_2$.

\subsection{Visualization (RQ5). \label{app:visual}}
We visualize node representations of the target domain generated by the baseline methods: GCN, DANE, UDAGCN, our proposed GCN-\model{} and its variant GCN-\model{}-IMOnly (without the Structure Consistency optimization objective).  
The aim is to further prove the performance of \model{} even without explicit domain adaptation component to mitigate the domain discrepancy.
One thing we want to point out is that the key to achieving good classification performance on the target domain is good class separability (the learned target node representations with the same label are close and the target node representations with different labels are far). 
Mitigating the domain discrepancy with an explicit domain adaptation component, which has been adopted by many UGDA methods, is just one of the effective approaches to enhance class separability on the target domain. However, it may not be necessary. The method like \model{} can also achieve good separation by both fully exploring the potential of the source model and utilizing the target graph structure.

For simplicity, we choose the most difficult task, DBLPv8 $\to$ ACMv9, to visualize the node representation.
The node representation is the hidden representation closest to the final linear full-connected layer, which is of the same dimension size in all methods: 128.
The dimensionality reduction method for visualization is the T-distributed Stochastic Neighbor Embedding (t-SNE)~\cite{van2008visualizing}. 
The visualization results of different methods are shown in Fig.~\ref{fig:visual1}. The color of each node represents its label.

We can observe that the well-trained source model GCN shows low-class separability on the target graph, where clusters have many overlaps. 
UGDA methods, DANE and UDAGCN, are somehow better with better clustering. However, the boundaries are still difficult to find. 
For our proposed GCN-\model{} marked in the red box, the large boundary can be seen though there are some nodes clustered mistakenly due to the limitation of the totally unsupervised adaptation procedure. 
We further compare the GCN-\model{} and its variant GCN-\model{}-IMOnly without SC optimization object to see how the structural information benefits the class separation.  
This phenomenon further reveals that with the graph structure to learn structure proximity, the source model can better adapt to the target data distribution.

%% file: src/conclusion.tex
\section{Conclusion}
In this work, we articulate a new scenario called Source Free Unsupervised Graph Domain Adaptation with no access to the source graph because of practical reasons like privacy policies.
Existing methods cannot work well as it is impossible for feature alignment. 
Facing challenges in SFUGDA, we propose our algorithm SOGA, which could be applied to arbitrary GNNs by adapting to the target domain distribution and enhancing the discriminative ability of the source model.
Extensive experiments indicate its effectiveness.

\section{Ethical Considerations}
In this study, we propose a novel OOD scenario with privacy preservation exploring methods to mitigate such OOD issues. Consequently, we do not foresee any obvious negative broader impacts in our paper. One only potential is that it remains unclear whether the algorithm could ensure fairness despite better privacy and safety.

%% file: src/appendix.tex
\setcounter{lem}{0}

\section{Theorem Proof \label{app:proof}}
\subsection{Proof of Lemma 1 \label{app:proof1}}
\begin{lem}
	  When the source model is optimized by the objective Eq. (4) with a gradient descent optimizer and the capacity of the source model is sufficiently large, for each node $v_i$ on the target graph, the predicted conditional distribution $q(y|x_i, \mathcal{N}_i; \Theta)$ will converge to a vector \textbf{q}, where the value of $\eta$ elements will be $\frac{1}{\eta}$, and the other elements will be $0$. $\eta$ is determined by the number of categories with the maximum probability value predicted by the original source model $q(y|x_i, \mathcal{N}_i; \Theta_s)$. Similarly, the non-zero positions of \textbf{q} are the indices of categories with the maximum probability value.
\end{lem}
\begin{proof}
By ignoring a constant factor $\frac{1}{|\mathbf{V_t}|}$, the objective Eq. (4) can be rewritten as:
\begin{equation}
    H\!(\mathbf{\hat{Y}_t} | \mathbf{V_t}\!)  \propto - \sum_{v_i \in \mathbf{V_t}} \sum_{y = 1}^k q(y|x_i, \mathbf{\mathcal{N}_i}; \Theta) \!\log q(y | x_i, \mathbf{\mathcal{N}_i}; \Theta).
\end{equation}

Considering a node $v_i$, we have the conditional entropy loss for a single observation $v_i$:
\begin{equation}
\begin{split}
    J_i &= -\sum_{y = 1}^k q(y|x_i, \mathbf{\mathcal{N}_i}; \Theta) \!\log q(y | x_i, \mathbf{\mathcal{N}_i}; \Theta)\\
    &= -\sum_{y = 1}^k q_i(y) \!\log q_i(y),
\end{split}
\end{equation}
where $q_i(y) \triangleq q(y | x_i, \mathbf{\mathcal{N}_i})$ for brevity.
Because the deep learning models for classification problems always have a normalization operation (e.g., Softmax) to make the final unconstrained representation be a probability form, $q_i(y)$ has the following form:
\begin{equation}
    q_i(y) =  \frac{z_y^{(i)}}{\sum_{j=1}^k z_j^{(i)}},
\end{equation}
where $\mathbf{z^{(i)}} = [z^{(i)}_1, ..., z^{(i)}_k]$ is the representation for node $v_i$ and it should satisfy the condition $\forall j \in [1, k], z^{(i)}_j \geq 0$ through a function ranging $[0, +\infty)$ such as $\exp$ in Softmax. The loss can be rewritten as:
\begin{equation}
    J_i = -\sum_{y = 1}^k \frac{z_y^{(i)}}{Z_i} \!\log \frac{z_y^{(i)}}{Z_i}
\end{equation}
where $Z_{i} = \sum_{j=1}^k z_j^{(i)}$ is the normalization factor. Since we will use gradient descent to optimize the objective function, we calculate the gradient of the objective for $\mathbf{z}$:
\begin{equation}
    \frac{\partial J_i}{\partial z_y^{(i)}} = - \frac{Z_i-z^{(i)}_y}{Z^2_i}\log \frac{z_y^{(i)}}{Z_i} + \sum^k_{j: j \neq y}\frac{z^{(i)}_j}{Z^2_i}\log \frac{z_j^{(i)}}{Z_i}. 
\end{equation}

We use Mathematical Induction in the following proof. Assuming $k = 2$ that means we only have 2 classes $\{1, 2\}$, we set $y = 1$ without loss of generality, and then we have:
\begin{equation}
\begin{split}
    \frac{\partial J_i}{\partial  z_1^{(i)}} &= - \frac{z^{(i)}_2}{Z^2_i}\log \frac{z_1^{(i)}}{Z_i} + \frac{z^{(i)}_2}{Z^2_i}\log \frac{z_2^{(i)}}{Z_i}\\
    &= \frac{z^{(i)}_2}{Z^2_i}\log \frac{z_2^{(i)}}{z_1^{(i)}}
\end{split}
\end{equation}
If $z_1^{(i)} = z_2^{(i)}$, then $\frac{\partial J_i}{\partial z_1^{(i)}} = 0$ and the objective is converged. The results will be $q_i(y=1) = q_i(y=2) = \frac{1}{2}$. If $z_1^{(i)} > z_2^{(i)}$, then $\frac{\partial J_i}{\partial z_1^{(i)}} < 0$ and $\frac{\partial J_i}{\partial z_2^{(i)}} > 0$. According to the gradient descent algorithm, $z_1^{(i)}$ will be larger, and $z_2^{(i)}$ will be smaller in the next iteration, which means that the order of $z_1^{(i)}$ and $z_2^{(i)}$ will not be changed. Because $z_2^{(i)} \geq 0$,  $q_i(y=2)$ will converge to 0 and $q_i(y=1)$ will converge to 1.
Finally, we have the conclusion that the prediction probability of a certain class will converge to 1 where this class has the largest prediction probability in the initial stage. If two classes have the same initial prediction probability, the probability will converge to $\frac{1}{2}$. Thus, the Lemma is proven for a binary classification problem.

We assume the Lemma is correct for a classification problem with $k-1$ classes. Let us see the problem with $k$ classes. If $z_1^{(i)} = z_2^{(i)} = ... = z_k^{(i)}$, then $q_i(y)$ will converge to $\frac{1}{k}$. If they are not all equal, there $\exists\; b, c \in [1, k]$ satisfy $b \neq c$ and $z_b^{(i)} > z_c^{(i)}$, and then we have
\begin{equation}
    \begin{split}
        \frac{\partial J_i}{\partial z_b^{(i)}} - \frac{\partial J_i}{\partial z_c^{(i)}} = \frac{1}{Z_i}\log \frac{z_c^{(i)}}{Z_i} - \frac{1}{Z_i}\log \frac{z_b^{(i)}}{Z_i} < 0,
    \end{split}
\end{equation}
which means that the gradient is monotonically decreasing for $z^{(i)}_j$. Thus, the order of $z^{(i)}$s will not be changed during the optimization procedure. 

If the class $m$ has the smallest $z_{m}^{(i)}$ and the class $m'$ has the second smallest $z_{m'}^{(i)}$, then we have:
\begin{equation}
    \begin{split}
        \frac{\partial J_i}{\partial z_m^{(i)}} &= - \frac{Z_i-z^{(i)}_m}{Z^2_i}\log \frac{z_m^{(i)}}{Z_i} + \sum^k_{j: j \neq m}\frac{z^{(i)}_j}{Z^2_i}\log \frac{z_j^{(i)}}{Z_i}\\
        &> - \frac{Z_i-z^{(i)}_m}{Z^2_i}\log \frac{z_m^{(i)}}{Z_i} + \sum^k_{j: j \neq m}\frac{z^{(i)}_j}{Z^2_i}\log \frac{z_{m'}^{(i)}}{Z_i}\\
        &= \frac{Z_i-z^{(i)}_m}{Z^2_i} \log \frac{z^{(i)}_{m'}}{z_m^{(i)}} > 0,
    \end{split}
\end{equation}
which means that the probability of the class with the smallest initial prediction probability will converge to 0. After a class converges to 0, the problem is equivalent to the classification problem with $k-1$ classes. Thus, we proved the Lemma for multiple classes. Because we have the assumption that the capacity of the source model is sufficiently large, the conclusion can be generalized to all the nodes. 
\end{proof}

\subsection{Proof of Lemma 2 \label{app:proof2}}
\begin{lem}
	  When the original source model is trained for a binary classification problem with the discriminative ability of $r_p$ and $r_n$ accuracy for positive samples and negative samples on the target graph respectively, the lower bound of AUC can be raised from $r_p \times r_n$ to $\frac{1}{2}(r_p + r_n)$ by using the conditional entropy objective Eq. (4). 
\end{lem}
\begin{proof}
Suppose we have $P$ positive nodes and $N$ negative nodes. We have $r_p$ and $r_n$ accuracy for positive and negative nodes, and we define the following variables:
\begin{equation}
    \begin{split}
        P_f \triangleq (1 - r_n) \times N, \quad P_t \triangleq r_p \times P, \\
        N_f \triangleq (1 - r_p) \times P, \quad N_t \triangleq r_n \times N, 
    \end{split}
\end{equation}
where $P_f$, $P_t$, $N_f$, and $N_t$ are the numbers of false positive, true positive, false negative, and true negative samples, respectively. If we do not conduct additional optimization, the worst-case AUC will be:
\begin{equation}
    {\it AUC}_l = \frac{P_t \times N_t}{N \times P} = r_1 \times r_2.
\end{equation}
Since the number of nodes whose initial prediction probabilities of two classes are exactly equal is rare, we ignore such nodes in the following proof for brevity. After adopting our optimization method, the prediction probability will be forced to 1 or 0 according to the initial prediction. Thus, the AUC can be calculated as follows:
\begin{equation}
\begin{split}
    {\it AUC} &= \frac{P_t \times N_t + \frac{1}{2}P_t \times N_f + \frac{1}{2}P_f \times N_t}{N \times P}\\ 
    &= \frac{1}{2}(r_1 + r_2).
\end{split}
\end{equation}
\end{proof}

\section{Experimental Details \label{app:experiment}}
In this section, we describe the hyperparameter, hardware, and software settings in detail.
The optimizer Adam \cite{kingma2014adam} with a learning rate of 0.01 is used for optimization. 
For methods including GNN, we use ReLU as the activation function and two-layer architectures where the hidden layer dimensions are [256, 128].
For methods applying negative sampling, we set the negative sampling number as 5.
For graph embedding methods, we take advantage of the open-source project: https://github.com/shenweichen/GraphEmbedding. Hyperparameters of the graph embedding methods are all the default settings. 
To ensure good performance on GraphSAGE and GAT, DANE, and UDAGCN, we use the grid search to find the optimal hyperparameters carefully. 
The number of GraphSAGE neighborhood samples of each layer is selected from [5, 10, 15, 20, 25, 30, 35, 40, 45, 50, 60, 70, 80, 90, 100, 150, 200, 250, 300, 350, 400, 450, 500]. 
The attention head of GAT is selected from [1, 2, 4, 8, 16, 32, 64]. 
When reproducing DANE, we conduct the grid search on the weight of the adversarial loss: $\lambda$, which corresponds to the LAMBDASINGLE in the open-source code of DANE. 
The value of $\lambda$ is selected among the integer numbers from 1 to 5.
When reproducing UDAGCN, we find that its performance is highly correlated with the length of the random walk, $l$, which is used to generate the PPMI matrix for the Global Consistency Network. 
A bad choice of $l$ may cause the failure of training.
The grid scope of $l$ is integer numbers from 1 to 10. 
We utilize the graph contrastive learning framework~\cite{zhu2021empirical} for tuning DGI and GRACE methods.
When reproducing GTrans, we utilize the same search space as the original paper. We search the learning rate of feature adaptation $\eta_1$ in [5e-3, 1e-3, 1e-4, 1e-5, 1e-6], the learning rate of structure adaptation $\eta_2$ in [0.5, 0.1, 0.01], the modification budget $B$ in [0.5\%, 1\%, 5\%] of the original edges.

In addition, for the hyperparameter-sensitive experiments, the different choices of $\lambda_1$ and $\lambda_2$ are: 
(1) for neighborhood evaluation, $\lambda_1$ is set to 1.0 and $\lambda_2$ is set to [0.1, 0.2, 0.3, 0.4, 0.5, 0.6, 0.7, 0.8, 0.9, 1.0] 
(2) for structure role evaluation, $\lambda_1$ is set to 1.0 and $\lambda_2$ is set to [0.1, 0.2, 0.3, 0.4, 0.5, 0.6, 0.7, 0.8, 0.9, 1.0]
For the hardware environment, all the experiments are performed on a Linux server (CPU: Intel(R) Xeon(R) CPU E5-2690 v4 @2.60GHz, GPU: NVIDIA Tesla K80s). 
For the software environment, python libraries for implementation are PyTorch 1.7.1 and torch-geometric 1.6.3.

\section{Dataset Details \label{app:dataset}}
Datasets ACM-D and ACM-S in this paper are the homogeneous parts selected from the heterogeneous citation graph data sets with only a paper-paper meta path.
The origin open source heterogeneous citation graph data set is collected from \cite{kong2012meta}.
They are available at https://github.com/PKUterran/MuSDAC.
More details about the collection of the origin datasets can be found in \cite{yang2020unsupervised}. 
Datasets DBLPv8 and ACMv9 in this paper are collected by \cite{wu2020unsupervised} from Aminer \cite{tang2008arnetminer}. They are available at https://github.com/GRAND-Lab/UDAGCN.